\pdfoutput=1

\documentclass[11pt]{article}

\usepackage[]{acl}

\usepackage{times}
\usepackage{latexsym}
\usepackage{url}
\usepackage{graphicx, subfigure, float}
\usepackage{mathrsfs}
\usepackage{amsfonts}
\usepackage{psfrag, amssymb, amsmath}
\usepackage{color,xcolor}
\usepackage{amsmath}
\usepackage{algorithm}
\usepackage{algorithmic}
\usepackage{wrapfig}

\usepackage[T1]{fontenc}

\usepackage[utf8]{inputenc}

\usepackage{microtype}

\title{Human-in-the-loop Robotic Grasping using BERT Scene Representation}

\author{
$\textbf{Yaoxian Song}^{1,2}$, 
$\textbf{Penglei Sun}^{1}$, 
$\textbf{Pengfei Fang}^{3}$, 
$\textbf{Linyi Yang}^{2,4}$, \\ 
$\textbf{Yanghua Xiao}^{1}$,
$\textbf{Yue Zhang}^{2,4}$\thanks{\ \ \ Corresponding Author} \\
$^{1}$School of Computer Science, Fudan University \\
$^{2}$School of Engineering, Westlake University \\
$^{3}$School of Engineering, Australian National University \\
$^{4}$ Institute of Advanced Technology, Westlake Institute for Advanced Study\\
\texttt{\{songyaoxian, yanglinyi, zhangyue\}@westlake.edu.cn}, \\
\texttt{\{plsun20, shawyh\}@fudan.edu.cn}, \texttt{u5765437@anu.edu.au}
}

\begin{document}
\maketitle
\begin{abstract}
Current NLP techniques have been greatly applied in different domains. In this paper, we propose a human-in-the-loop framework for robotic grasping in cluttered scenes, investigating a language interface to the grasping process, which allows the user to intervene by natural language commands. This framework is constructed on a state-of-the-art grasping baseline, where we substitute a scene-graph representation with a text representation of the scene using BERT. Experiments on both simulation and physical robot show that the proposed method outperforms conventional object-agnostic and scene-graph based methods in the literature. In addition, we find that with human intervention, performance can be significantly improved. Our dataset and code are available on our project website\footnote{\url{https://sites.google.com/view/hitl-grasping-bert}}.
\end{abstract}


\section{Introduction}



\begin{figure}[t]
	\centering
	\setlength{\belowcaptionskip}{-0.5cm} 
	\includegraphics[width=\linewidth]{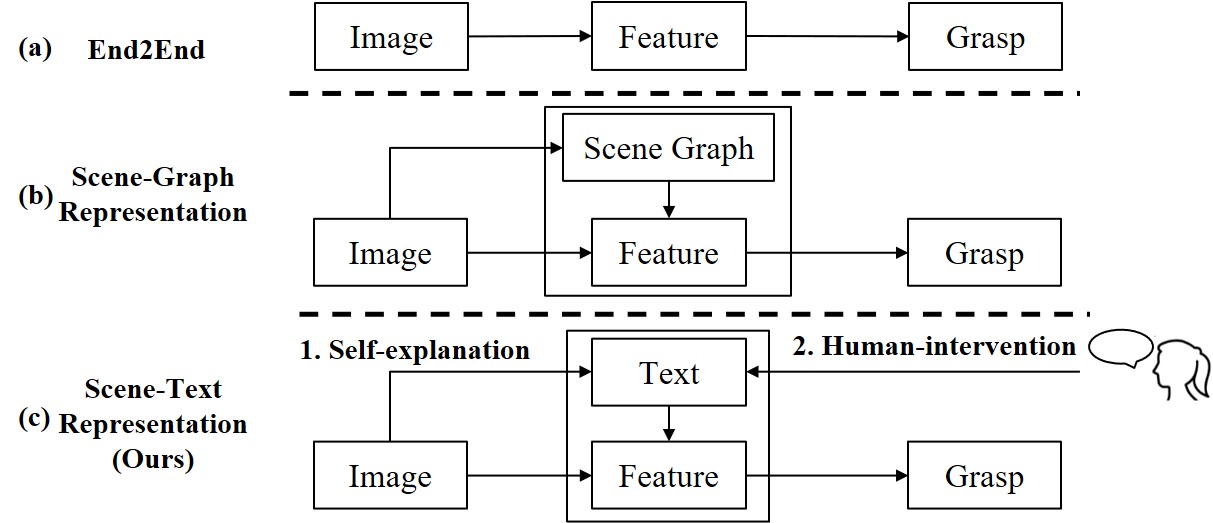}
	\caption{Various model structures for robotic grasping. (a) \textbf{End2End} \cite{chu2018real} outputs an object-agnostic grasping by directly modeling on the input images. (b) \textbf{Scene-Graph Representation} fuses a generated scene graph with visual feature to predict grasping. (c) \textbf{Scene-Text Representation (ours)} makes use of language scene description and visual feature for achieving collision-free grasping.}
	\label{fig:model_conf}
\end{figure}
\begin{figure}[t]
	\centering 
	\setlength{\abovecaptionskip}{-0.1cm}  
    \setlength{\belowcaptionskip}{-0.5cm} 
	\subfigure[Overview of proposed dataset L-VMRD]{
	\includegraphics[width=0.98\linewidth]{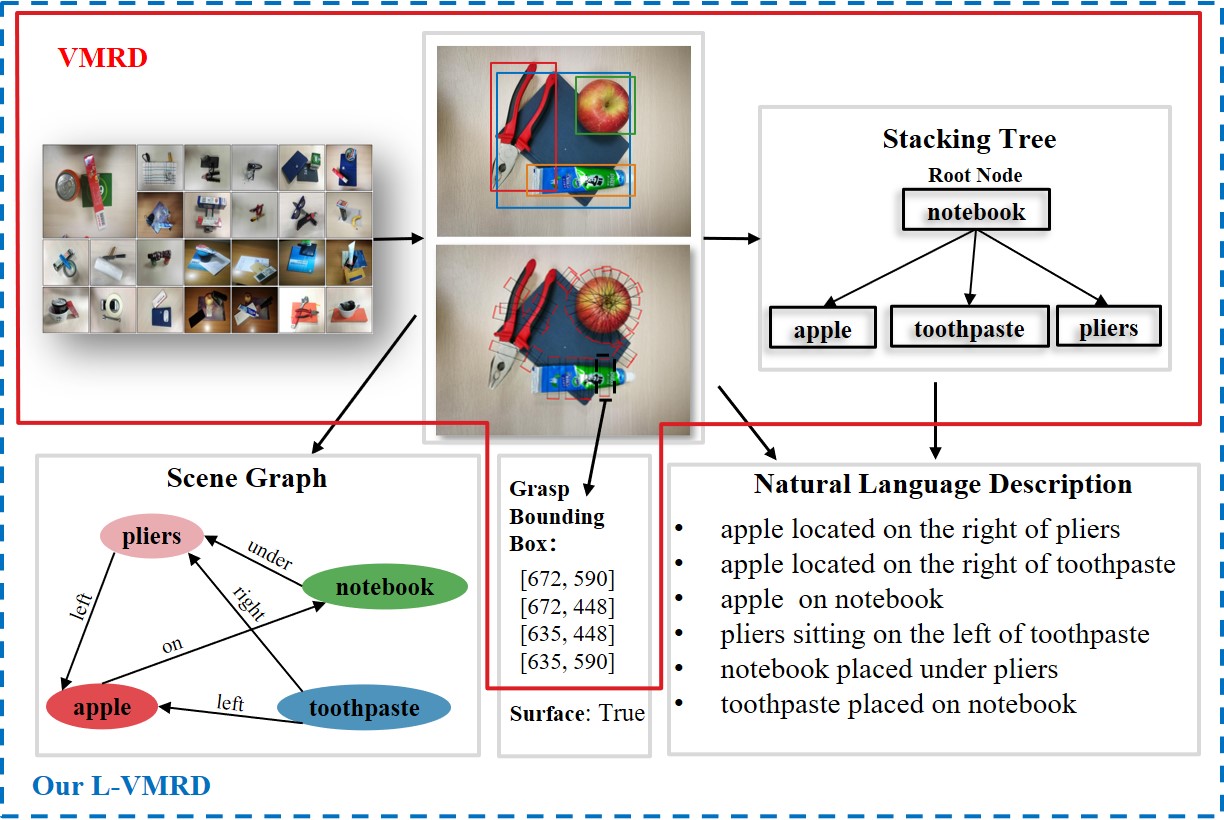}
	\label{fig:dataset}
}
\\
	\subfigure[Relationship tree \& Scene graph.]{
	\includegraphics[width=0.98\linewidth]{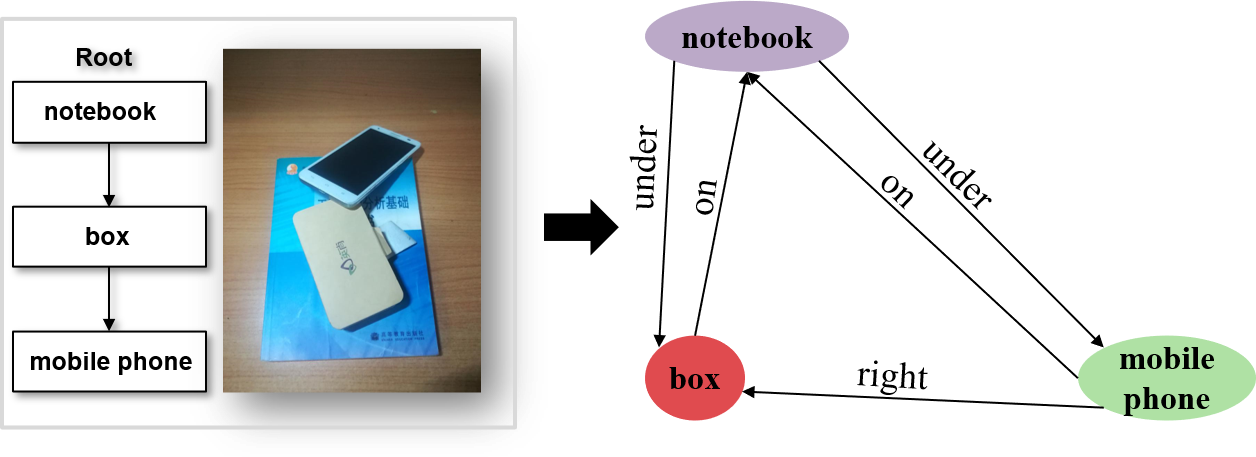}
	\label{fig:dataset_sg}
	}
	\caption{\textbf{(a):} L-VMRD is built on VMRD. We extend \textbf{(i)} scene language description, \textbf{(ii)} scene graph and \textbf{(iii)} \textbf{surface} per grasp, including $112,965$ scene object relationship expressions and $21,713$ \textbf{surface} attributes paired with grasp bounding boxes. \textbf{(b):} relationship tree vs. scene graph.}
	\label{fig:dataset_all}
\end{figure}

Grasping \cite{mahler2019learning} is a fundamental task for robot systems. It is useful for warehousing, manufacturing, medicine, retail, and service robots. 
One setting in robotic grasping is to grasp object orderly without disturbing the remaining in cluttered scenes \cite{chen2021joint,mees2020composing,zhang2021invigorate} (called \textbf{collision-free grasp}). To solve this problem, a typical method \cite{zhu2020hierarchical} parses the input into a \textbf{scene graph} first (Figure.~\ref{fig:model_conf}(b)), in order to infer the spatial relation between objects and select a collision-free object for grasping. In particular, as shown in Figure.~\ref{fig:dataset}, nodes in a scene graph represent objects and edges represent their spatial relationship. Such structural knowledge can effectively improve the grasping performance as compared to an end-to-end model without scene structure information \cite{chu2018real} (Figure.~\ref{fig:model_conf}(a)).

As a structured representation of the scene, the scene graph has a few limitations. For example, it can be costly to manually label scene graphs, and the amount of existing labeled data is quite small. In practice, due to variance in the working environment, it can be necessary to calibrate a scene understanding model when deployed on physical robots \cite{zhu2020hierarchical}. This problem can be regarded as a domain adaptation task, which requires a certain amount of labeled scene graph data \cite{xu2017scene}. In addition, graphs are relatively abstract and thus inconvenient for human-robot interaction. 



We consider a natural language representation of the scene for substituting a scene graph structure representation. As shown in Figure.~\ref{fig:dataset}, small texts such as \textit{``notebook placed under pliers''} and \textit{``apple on notebook''} are used to indicate the recognized objects and their stacking relations. Compared with a scene graph structure, natural language scene representation has the following advantages. First, the cost of manual labeling is relatively lower thanks to the availability of speech recognition systems \cite{chiu2018monotonic} and the relative independence from labeling GUI \cite{srivastava2021behavior}. Second, the state-of-the-art pre-trained representation models \cite{kenton2019bert,radford2019language} can be used to improve scene understanding, which contains external knowledge beyond a scene graph structure. Third, online human interaction can 
be achieved by using human input of the natural language scene representation to replace incorrect robot scene understanding through speech communication\footnote{We adopt the \textbf{typed text} to simulate the process here since voice recognition is beyond our research scope}.

As shown in Figure.~\ref{fig:model_conf}(c), we adopt the model of \citet{zhu2020hierarchical} by substituting the scene graph with a text description of the object to grasp and its spatial context, and using a neural image-to-text model for scene understanding and a pre-trained language model to represent the scene text for visual language grounding in subsequent grasping decisions. We compare the model performance with a dominant two-stage end-to-end planar grasping baseline \cite{chu2018real} (Figure.~\ref{fig:model_conf}(a)) and the baseline scene graph model (Figure.~\ref{fig:model_conf}(b)). For all models, the grasping backbone is implemented using an extended version model of \cite{chu2018real} with extra scene knowledge input. 

For training and evaluation, we make extensions to the Visual Manipulation Relationship Dataset (VMRD) \cite{zhang2019roi} by manually adding text descriptions and scene graphs to the scenes, resulting in a new dataset \textbf{L-VMRD}, as shown in Figure.~\ref{fig:dataset}. Experimental results show that (1) human language description can be a highly competitive alternative to the scene graph representation, giving better results for grasping; (2) online human language intervention is useful for improving the final grasping results, which is a new form of human-in-the-loop grasping. This indicates the promise of NLP models, especially pre-trained language models, for human-robot interaction. 
To our knowledge, we are the first to consider explicit textual scene representation and human intervention correction for robot grasping decisions, where BERT~\cite{kenton2019bert} is firstly introduced into the internal structure of a robotic model as a state representation. 




\begin{figure*}[t]
	\centering 
	\setlength{\belowcaptionskip}{-0.5cm}
	\includegraphics[width=0.9\linewidth]{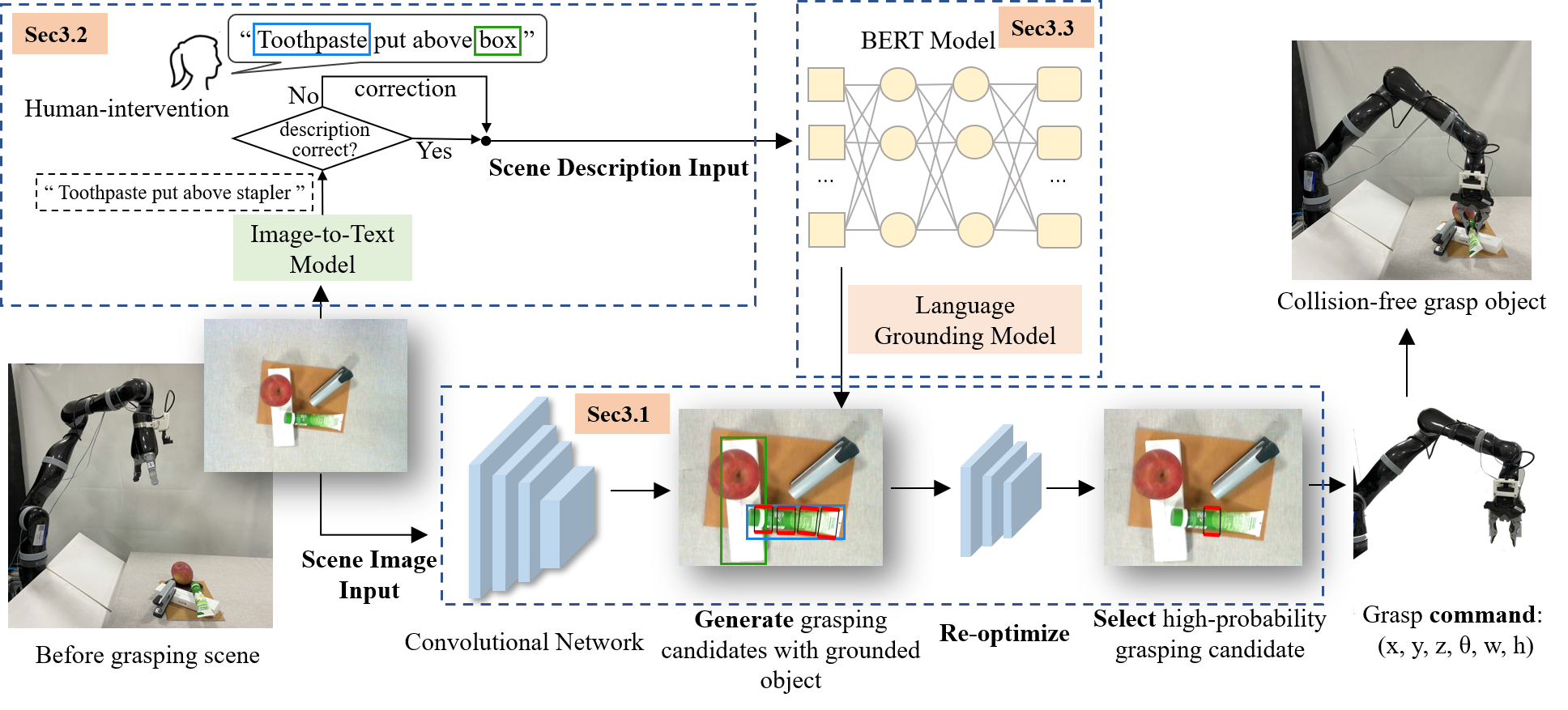}
	\caption{Overview of our proposed model. Firstly, a scene image is given to a human and an image-to-text model to generate a scene description. Secondly, scene description is fed into a BERT-based language grounding model to select a collision-free object. Thirdly, the grounded object is as the internal result fused into the language-based grasping model.}
	\label{fig:network_struture}
\end{figure*}

\section{Task and Dataset}\label{sec:task_dataset}
The input of the robotic grasping task is an image from robotic camera observation and the output is a grasping configuration (a grasping bounding box). As shown in Figure.\ref{fig:model_conf}(c), we introduce a language description for the scene (image) during the inference process. We take a human-in-the-loop setting, where the language description can be obtained from a scene understanding model (\textbf{Self-explanation}) or human (\textbf{Human-intervention}). 

Existing grasping datasets (e.g., VMRD~\citet{zhang2019roi}) cannot be employed directly, because they do not include scene knowledge in human language. Hence, we develop an extended language version of VMRD, named L-VMRD, to evaluate our method. L-VMRD is an integrated dataset, and each sample is organized as a 6-tuple \textbf{(image, language descriptions, scene graph, object bounding box, grasping bounding box, surface)} shown in Figure.~\ref{fig:dataset_all}. L-VMRD contains $4,676$ samples, and is split into (train/validate/test) $=(3,740/468/468)$. The detail of dataset generation and usage in modeling are demonstrated in Appendix~\ref{app:dataset_all} and \ref{app:dataset_use}. Below we describe the main extensions from VMRD (Figure.\ref{fig:dataset}).


\textbf{Language Description in L-VMRD}
Object pairs in an image are sampled and labeled with a scene description. There exist many factors that affect collision-free grasping, in which stacking is the most significant one \cite{avigal2021avplug}. We first label the objects with the stacking relationship, e.g., \textit{``apple on notebook''} or \textit{``notebook sitting under pliers''} in Figure.~\ref{fig:dataset}, and then label the horizontal relationship between non-stacked. Then considering the distribution of the horizontal relationships (\textit{``left''}, \textit{``right''}, \textit{``front''}, \textit{``back''}) in our dataset, we use \textbf{\textit{``left''}} and \textbf{\textit{``right''}} to indicate the scene relationships. 

\textbf{Scene Graph}
The original VMRD dataset includes partial relationships for adjacent objects, encoded by a relationship tree that only reveals the stacking relationship between two objects but is not able to indicate the relationship between objects stacked directly. To facilitate inference, we add a full scene graph to encode the pair-wise relationships of objects per image, where nodes and edges present objects and relationships,  respectively. Detail is available in Appendix~\ref{app:sg}.

\textbf{Grasp-wise Spatial Attribute}
We introduce each grasp bounding box with a new attribute named \textbf{surface}. It is a binary variable indicating whether the grasped object sits on the top (True -- on the top, False -- stacked by other objects.). It is a grasp-wise label and can improve the robustness of the grasping model. An example is available in Appendix~\ref{app:surface}.


\section{Our Approach}
The overall framework is shown in Figure.~\ref{fig:network_struture}. The input scene image is fed into convolutional-based grasping model (Sec.~\ref{sec:grasping_model}) and scene understanding model (Sec.~\ref{sec:self-exp}; Sec.~\ref{sec:lang_grounding}), respectively. The scene understanding model is an image-to-text component that produces a sentence that describes the object to grasp and its context, such as \textit{``toothpaste put above box''}. That scene description is then fused with intermediate grasping results to select object-related grasping candidates by pre-trained language model and language grounding model. \textbf{The final grasping output} is selected with high probability after re-optimization. A grasp command is sent to a real robot to complete a collision-free grasping operation. For the human-in-the-loop scenario, \textbf{an extra conditional input} from human-intervention will be given to correct scene description when the description from the image-to-text model is incorrect. 



\subsection{Overall of Language-based Grasping Model}\label{sec:grasping_model}
Let $I$ denote an image as perceptive information from the environment (i.e., cluttered scenes). Our robot  $f$ first identifies the grasp configuration from observation $I$. A typical 5-dimensional grasp configuration is given by:
\begin{equation}\label{eq:grasp_def1}
\small
	g_i = f(I) = (x,y,\theta, w, h),
\end{equation}
where $(x, y)$ is the center position of the grasp rectangle, $\theta$ is the orientation angle with the x-axis, and $(w, h)$ is the weight and height of the grasp rectangle. A general robotic grasping is presented by a probability $P(g_i|I)$, where $g_i \in G$ and $G$ is a set of grasping candidates. 

To achieve more stable and safe grasping, a joint probability $P(g_i, K_g|I)$ can integrate additional scene knowledge $K_g$ as auxiliary information to guide vision-based grasping. It can be decomposed into conditionally independent two parts, given by:
\begin{equation}\label{eq:model}
\small
P\left( {{g_{_i}, K_g}|I} \right) = P\left( {{g_{_i}}|I, K_g} \right)P\left({K_g|I} \right),
\end{equation}
where $P(K_g|I)$ is a scene understanding model. It can be a scene structure parsing model (\citealp{zhu2020hierarchical}; Figure.~\ref{fig:model_conf}(b)), an image-to-text model (Figure.~\ref{fig:model_conf}(c); Sec.\ref{sec:self-exp}) with grounding model (Sec.\ref{sec:lang_grounding}) or direct human intervention with grounding model (Sec.\ref{sec:lang_grounding}).  $P\left( {{g_{_i}}|I, K_g} \right)$ is a convolutional network and the details are described in Appendix~\ref{app:grasping_model}.


\subsection{Grasping Scene Understanding}\label{sec:self-exp}
A state-of-the-art image-to-text model (MMT) \cite{cornia2020meshed} is used to generate the scene description in our work, which is a standard encoder-decoder Transformer-based model \cite{vaswani2017attention}, that learns a multi-level representation of the relationships between image regions integrating with learned prior knowledge, and uses a mesh-like connectivity at decoding stage to exploit low- and high-level features. More details are in Appendix~\ref{app:image-to-text}.

\textbf{Encoder} A set of image regions $I$ as Input is fed into encoding layer following Eq.~\eqref{eq:encoder}.
\begin{equation}\label{eq:encoder}
\small
\begin{aligned}
& Z = AddNorm\left( {{M_{mem}}\left( I \right)} \right),\\
& \tilde X = AddNorm\left( {F\left( Z \right)} \right),
\end{aligned}
\end{equation}
where $AddNorm$ indicates the composition of a residual connection and layer normalization. $M_{mem}$ is memory-augmented attention operation in Eq.~\eqref{eq:mem}. $F\left( Z \right)$ is a position-wise feed-forward layer composed of two affine transformations with a single non-linearity. $\tilde X=(\tilde X^1,...,\tilde X^i...,\tilde X^N)$ is the set of all encoding layers and $N$ is the number of layers.
\begin{equation}\label{eq:mem}
\small
\begin{aligned}
& {M_{en}}\left( I \right) = Attention\left( {{W_q}I,K,V} \right),\\
& K = \left[ {{W_k}I,{M_k}} \right],\\
& V = \left[ {{W_v}I,{M_v}} \right],
\end{aligned}
\end{equation}
where $Attention$ is the self-attention operations used in \cite{vaswani2017attention}. $W_q$, $W_k$, $W_v$ are matrices of learnable weights. $M_k$ and $M_v$ are learnable prior information. 


\textbf{Decoder} The decoder takes an input sequence of vector $Y$ and output layers from encoder $\tilde X$, and then outputs sequence $\tilde Y$, in Eq.~\ref{eq:de}.
\begin{equation}\label{eq:de}
\small
\begin{aligned}
& Z = AddNorm\left( {{M_{de}}\left( {\tilde X} \right),Y} \right),\\
& \tilde Y = AddNorm\left( {F\left( Z \right)} \right),\\
\end{aligned}
\end{equation}
where $M_{de}$ is defined in Eq.~\ref{eq:de_detail}. $Y$ is the input sequence of vector (groundtruth).
\begin{equation}\label{eq:de_detail}
\small
\begin{aligned}
& {M_{de}}\left( {\tilde X,Y} \right) = \sum\limits_{i = 1}^N {{\alpha _i} \odot } C\left( {{{\tilde X}^i},Y} \right),\\
& C\left( {{{\tilde X}^i},Y} \right) = Attention\left( {{W_q}Y,{W_k}{{\tilde X}^i},{W_v}{{\tilde X}^i}} \right),\\
& {\alpha _i} = \sigma \left( {{W_i}\left[ {Y,C\left( {{{\tilde X}^i},Y} \right)} \right] + {b_i}} \right),
\end{aligned}
\end{equation}
where $C$ is cross-attention operation, $\sigma$ the sigmoid activation function, and $\odot$ element product.

We make use of two adaptations during training. The first replaces the region features in MMT\footnote{https://github.com/aimagelab/meshed-memory-transformer}  \cite{cornia2020meshed} with the concatenation of region features with bounding box features. Secondly, we add an extra score to multiply the CIDEr-D reward \cite{rennie2017self} during training by maximizing a reinforcement learning based reward, since the description of subject object is usually the grasped one in our task. The score is computed by the correct rate of the subject over all generated sentences for each training batch.



\subsection{Language Grounding for Grasping}\label{sec:lang_grounding}
We make use of visual language grounding models to map a scene description to a specified object. 
For visual language grounding, let $Q$ represent a query sentence from human or image-to-text model and $I \in R{^{H \times W \times 3}}$ denote the image of width $W$ and height $H$. The task aims to find the object region $K_g$ represented by its center point $(x_t, y_t)$ and the object size $(w_t, h_t)$. The overall method can be formulated as a mapping function $(x_t, y_t, w_t, h_t)=\phi(Q, I)$.

In this paper, considering the real-time robotic control, we deploy our task on a one-stage language grounding model\footnote{https://github.com/zyang-ur/onestage\_grounding} \cite{yang2019fast} based on YOLOv3\footnote{YOLOv3 is a typical object detection model and derives many multimodal variants.} \cite{redmon2018yolov3} with different language encoders for the mapping function $\phi$. 
Formally, the scene image $I$ and scene description $Q$ are input to the visual encoder and text encoder, respectively, and the grounding module outputs the grounded object with encoders' features following Eq.~\ref{eq:grounding}.
\begin{equation}\label{eq:grounding}
\small
\begin{aligned}
& {Z_{vis}} = {M_{vis}}\left( I \right),\\
& {Z_{lang}} = {M_{lang}}\left( Q \right),\\
& K_g = Ground\left( {\left[ {{Z_{vis}},{Z_{lang}},{Z_{spatial}}} \right]} \right),
\end{aligned}
\end{equation}
where $M_{vis}$ is Darknet-53 \cite{redmon2018yolov3} pre-trained on COCO object dataset \cite{lin2014microsoft} and fine-tuned on our proposed L-VMRD. $Ground$ is the same as the output layers of YOLOv3. $Z_{spatial}$ is spatial feature of visual feature defined as follows:
{$(\frac{i}{W},\frac{j}{H},\frac{i+0.5}{W},\frac{j+0.5}{H},\frac{i+1}{W},\frac{j+1}{H},\frac{1}{W},\frac{1}{H})$,}
which denotes top-left, center, bottom-right plane coordinates, sizes of the each pixel in visual feature mapping $Z_{vis}$, respectively, normalized by the width $W$ and height $H$ of the feature mapping. $i \in \{0,1,...,W-1\}$ and $j \in \{0,1,...,H-1\}$.

For $M_{lang}$ of the text encoder, we choose BERT \cite{kenton2019bert}\footnote{We use bert-base-uncased model in our work.} and the encoder of Transformer \cite{vaswani2017attention} (simply named Transformer). BERT is a pre-trained language model and builds on the Transformer network. Each description is fed into $M_{lang}$, resulting in $768$ dimensions embeddings of all the tokens as natural language representations. Transformer\footnote{https://github.com/pytorch/examples/tree/master/\par word\_language\_model (\textbf{6 encoder\_layers implemented})} can be regarded as randomly initialized BERT without pre-training.
Each description is embedded into $1,024$ dimensions embeddings. The model is randomly initialized.

\subsection{Training Details} 
We break an end-to-end grasping training process into three submodules, \textbf{(i)} image-to-text (\textbf{self-explanation}), \textbf{(ii)} language grounding, and \textbf{(iii)} language-based grasping successively. The first two models are trained based on the original project configurations. The detail of the last one is described in Appendix ~\ref{app:grasp_training}.

\section{Evaluation}
We construct both simulation and physical experiments to investigate four research questions:


\noindent \textbf{Q1:} How much natural language (\textbf{unstructured}) scene description perform better than scene graph (\textbf{structured}) knowledge in collision-free grasping task?

\noindent \textbf{Q2:} How much does the pre-trained model perform better than the randomly initialized model in our task?

\noindent \textbf{Q3:} How and where does human intervention using NLP improve grasping performance under the proposed human-in-the-loop framework? 

\noindent \textbf{Q4:} How does our method perform on a physical robot? Does data collection from our proposed human-in-the-loop framework improve efficiency during the fine-tuning process?


\subsection{Settings}

\noindent\textbf{Scene and Platform Implementation} In simulation experiments, we use the test set split in Sec.~\ref{sec:task_dataset} to evaluate our method. In the physical experiment, we collect objects and set up the placement as similar as possible to the training set, in which 2-6 objects are randomly placed (stacked) on a white table. But the reality gap is usually inevitable, especially in a robotic environment, which also tests the robustness of our method implicitly. Other details of the robot and training framework are described in Appendix ~\ref{app:hard_soft}.


\begin{table*}[t]
\centering
\small
\scalebox{0.95}{
\begin{tabular}{lcccccccc}
\hline
\multicolumn{1}{c}{Method}                        & R@1  & R@3  & R@5  & R@10                      & P@1  & P@3  & P@5  & P@10 \\ \hline
\multicolumn{1}{l}{End2End}                      & 42.5          & 66.2          & 78.9          & \multicolumn{1}{c}{88.9}          & 42.5          & 40.5          & 40.1          & 37.5          \\
\multicolumn{1}{l}{SceneGraph-Rep}               & 72.2          & 85.6          & 90.3          & \multicolumn{1}{c}{92.2}          & 73.8          & 70.7          & 69.0          & 64.0          \\ \hline
\multicolumn{1}{l}{SceneText-Transformer}        & 72.2          & 85.6          & 88.3          & \multicolumn{1}{c}{90.3}          & 72.4          & 68.8          & 66.2          & 61.6          \\
\multicolumn{1}{l}{SceneText-BERT}               & \textbf{73.7} & \textbf{88.9} & \textbf{91.8} & \multicolumn{1}{c}{\textbf{93.1}} & \textbf{73.9} & \textbf{71.9} & \textbf{69.2} & \textbf{65.3} \\ \hline
\multicolumn{9}{c}{With Human-intervention}                                                                                                                                                            \\ \hline
\multicolumn{1}{l}{SceneText-Interv-Oracle}      & 77.0          & 89.1          & 90.8          & \multicolumn{1}{c}{91.4}          & 78.0          & 76.4          & 75.0          & 71.7          \\ 
\multicolumn{1}{l}{SceneText-Interv-Transformer} & 75.9          & 88.8          & 91.8          & \multicolumn{1}{c}{92.4}          & 76.3          & 71.2          & 69.2          & 64.9          \\
\multicolumn{1}{l}{SceneText-Interv-BERT}        & \textbf{76.9} & \textbf{90.3} & \textbf{93.0} & \multicolumn{1}{c}{\textbf{94.9}} & \textbf{76.3} & \textbf{75.3} & \textbf{73.6} & \textbf{69.1} \\ \hline
\end{tabular}} 
\caption{Results of self-explanation and human-intervention models in the simulation experiment. The best performance is highlighted in bold.}
\label{table:exp1}
\end{table*}

\noindent\textbf{Evaluation Metrics} We take Benchmark Performance and Success Rate as the main evaluation metrics. The first is used in both simulation and physical experiments, and the second one is only used in the physical robot experiment:

\noindent{\textit{$\bullet$ Benchmark Performance}:} In simulation experiment, we evaluate model performances of the collision-free grasp using the object retrieval \textit{top-k recall (R@k)} and \textit{top-k precision (P@k)} metrics to evaluate multi-grasp detection \cite{hu2016natural}. \citet{chen2021joint} proposes above metric to evaluate language-based multi-grasping. We do not compare it with our work directly, because: (i) their work (including dataset) is not open-sourced. (ii) it is just a command-based end-to-end grasping method that did not consider language scene understanding with human-in-the-loop. In physical robot experiment, \textbf{\textit{accuracy}} is the percentage of correct cases over all test cases. The correct case is defined in Appendix~\ref{app:metric}.

\noindent{\textit{$\bullet$ Success Rate}:} In the physical robot experiment, we calculate the percentage of successful collision-free grasps over all grasping trials.


\subsection{Models}\label{sec:model_define}
We compare different pipeline methods visualized in Figure.~\ref{fig:model_conf}. The baselines include:

\noindent\textbf{End2End} We re-train a state-of-the-art end-to-end object-agnostic planar grasp detection model MultiGrasp \cite{chu2018real} on L-VMRD, shown in Figure.~\ref{fig:model_conf}(a).

\noindent\textbf{SceneGraph-Rep} This is shown in Figure.~\ref{fig:model_conf}(b) using a structured form of the scene graph generation\footnote{https://github.com/jwyang/graph-rcnn.pytorch} (IMP) \cite{xu2017scene} encoded with relational graph convolution network (RGCN) \cite{schlichtkrull2018modeling} shown in Figure.~\ref{fig:network_struture}. It replaces the subprocess from Image-to-Text Model to Language Grounding Model in Figure.~\ref{fig:network_struture} to select the grouned object. See details in Appendix~\ref{app:sg_model}.

\noindent Our proposed models (\textbf{Scene-Text Representation} in Figure.~\ref{fig:model_conf} (c)) include:

\noindent\textbf{SceneText-\{BERT, Transformer\}} They are models using image-to-text MMT \cite{cornia2020meshed} with language grounding \cite{yang2019fast} to realize explainable grasping (\textbf{Self-explanation w/o Human-intervention}).  

\noindent\textbf{SceneText-Interv-\{BERT, Transformer\}} They bring in human-intervention shown in Figure.~\ref{fig:network_struture}.  In \textbf{SceneText-Interv-Oracle}, the retrieval region of language grounding from the groundtruth is fed directly into the downstream grasp model (\textbf{Self-explanation w/ Human-intervention}).


\begin{figure}[t]
	\centering 
	\setlength{\belowcaptionskip}{-0.5cm} 
	\includegraphics[width=\linewidth]{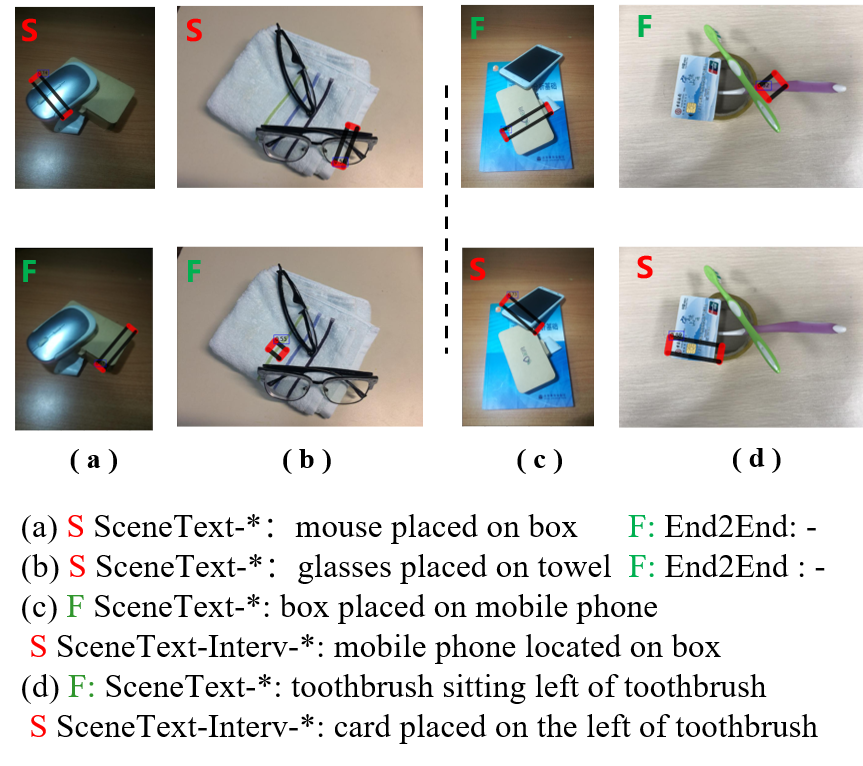}
	\caption{Visualization of baseline models and our proposed models in our work. \textcolor{red}{\textbf{S}} means a successful case, while \textcolor{green}{\textbf{F}} means a failure case.}
	\label{fig:case_study1}
\end{figure}

\subsection{Simulation Results}
\textbf{Scene Knowledge} Table~\ref{table:exp1} gives the results of the end-to-end, scene-graph based method, and our method on L-VMRD data. In this setting, the input of the model is only the image and models predict a collision-free grasping. Although End2End \cite{chu2018real} is the state-of-the-art method in object-agnostic grasping, it performs poorly on the cluttered scene grasping tasks. In contrast, by first obtaining a scene graph and then predicting the selected object, the SceneGraph-Rep model gives much improvement, with $72.2\%$ over End2End R@1. This shows the necessity of scene understanding for the collision-free grasping task. As a case study, in Figure.~\ref{fig:case_study1}(a)(b), SceneText-* generate self-explanation expression and obtain correct collision-free grasping, while End2End predicts incorrect grasping (incorrect object selection and low-quality grasp detection).


\noindent\textbf{Scene Graph vs. Natural Language} For \textbf{Q1}, compared with SceneGraph-Rep model which has R@1 of $72.2\%$, our method SceneText-BERT gives a better R@1 of $73.7\%$. This shows the feasibility of using natural language to replace the scene graph for object selection. Both methods are trained under the same settings, yet a natural language is more useful for achieving expandability in real-time human-robot interaction, thanks to its direct connection to the natural representation of the scene. For \textbf{Q2}, among our models, BERT can better parse the generated scene descriptions than Transformer model in both P@k and R@k, which shows the benefit of external pre-training. Note that Transformer alone does not outperform SceneGraph-Rep in Table~\ref{table:exp1}. The results show that pre-training allows a textual representation of the scene to compete with a standard graph representation. More case studies can be found in Appendix~\ref{app:sg_case}.


\noindent\textbf{Human-intervention} For \textbf{Q3},  the last two rows in Table~\ref{table:exp1} show that the models can take human language descriptions about the cluttered scene as a guidance or error correction for its own textual scene representation. By comparing results in Table~\ref{table:exp1}, we can find that our proposed human-in-the-loop framework improves the performance from SceneText-* to SceneText-Interv-*. For example, compared with SceneText-BERT, SceneText-Interv-BERT improves the R@1 value from $73.7\%$ to $76.9\%$, which shows that human intervention can be useful in practical scenarios. As a case study shown in Figure.~\ref{fig:case_study1}(c)(d), the image-to-text model generates an incorrect relationship between \textit{``box''} and \textit{``mobile phone''}, leading to a failed collision-free grasping detection on the stacked box. In contrast, an extra human scene language description corrects the self-explanation description and helps the model to obtain a correct collision-free grasping detection. Within the above process, $21.1\%$ scene language description from a robot is incorrect and intervened by human-correction. We also present the results of different human-intervention rates on F1 score, between SceneText-BERT and SceneText-Interv-BERT. As shown in Figure.~\ref{fig:dev}, with the increase of human-intervention, the F1 score improves steadily. This shows that human language intervention can compensate for the flaws of scene understanding from the image-to-text models effectively.

\begin{table}[t]
\centering
\scalebox{0.75}{
\begin{tabular}{lcccccc}
\hline
\multicolumn{1}{c}{Inter(\%)} & 0 & 25 & 50 & 75 & 100                       & \begin{tabular}[c]{@{}c@{}}baseline\\ (L-VMRD)\end{tabular} \\ \hline
\multicolumn{1}{l}{\textbf{MMT}}      & 44.6       & 45.4        & 49.2        & 51.7        & \multicolumn{1}{c}{\textbf{55.0}} & 12.5                                                                 \\ 
\multicolumn{1}{l}{\textbf{LangGr}}   & 91.3       & 91.7        & 90.8        & 92.1        & \multicolumn{1}{c}{\textbf{92.9}} & 79.6                                                                 \\ \hline
\end{tabular}}
\caption{Results of image-to-text accuracy and language grounding accuracy @0.5 in the physical experiment. \textbf{Inter-} for short of intervention rate, e.g., \textbf{Inter-100}.}
\label{table:exp2}
\end{table}

\subsection{Physical Robot Results}
\begin{figure}[t]
	\centering 
	\setlength{\belowcaptionskip}{-0.5cm} 
	\includegraphics[width=\linewidth]{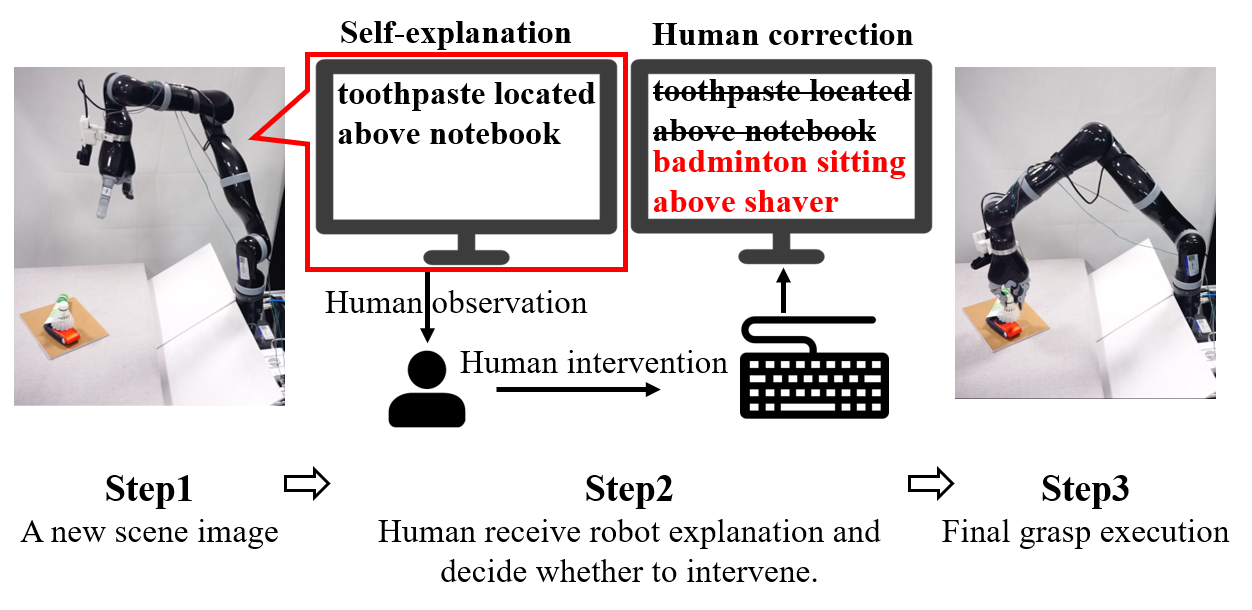}
	\caption{A pipeline of real robot execution with human.}
	\label{fig:exp_robot}
\end{figure}

We only investigate the performance of BERT-based models (SceneText-BERT, SceneText-Interv-BERT) in the physical robot experiment.
We recruit five graduate students to participate in our real-world experiment, who observe one whole process of a cluttered grasping with human-like grasping object selection each time by turns. When the image-to-text (self-explanation) model outputs an erroneous description, they correct the scene description by text typing decisively. Ultimately we collect $400$ samples, including $160$ correct output samples without human-intervention (self-explanation) and $240$ human corrected (intervention) samples using the model trained on L-VMRD. Each sample contains an image, a language description sentence, and a grounded object bounding box. A pipeline case containing human-intervention is shown in Figure.~\ref{fig:exp_robot}.

\noindent\textbf{Human-in-the-loop Learning Deployment}
Inspired by \citet{lu2022rationale}, we take $160$ samples as the training set to fine-tune our model based on different intervention rates (proportion of human-intervention samples, \textbf{Inter} for short). We randomly sample $80$ samples from the remaining $240$ human-intervention samples (getting rid of training used) as our test set. We repeat the process three times to generate three different test sets. The results in Table~\ref{table:exp2} show the mean values of models test on three test sets. 
For \textbf{Q4}, Table~\ref{table:exp2} shows the Benchmark Performance of models fine-tuned on different intervention rate training sets. 
The values of baseline are from the image-to-text model (\textbf{MMT}) and language grounding model (\textbf{LangGr}) trained on L-VMRD, respectively, which show the necessity of domain adaptation. For both MMT and LangGr, the models fine-tuned on human-intervention data (intervention rate $100\%$) achieve the best performance ($55.0\%$ and $92.9\%$). 
For MMT, the best model improves up to $\mathbf{42.5\%}$ compared to baseline ($12.5\%$) and $\mathbf{10.4\%}$ compared to the model fine-tuned on self-explanation ($44.6\%$, intervention rate $0\%$). For LangGr, the best model improves up to $\mathbf{13.3\%}$ compared to baseline ($79.6\%$) and $\mathbf{1.6\%}$ compared to the model fine-tuned on self-explanation data ($91.3\%$). This shows that the data collected from human-intervention can achieve better performance in domain adaptation from simulation to physical environment.

\begin{figure}[t]
	\centering 
    \setlength{\abovecaptionskip}{-0.1cm}  
    \setlength{\belowcaptionskip}{-0.5cm} 
    \subfigure[F1 score on different intervention rates.]{
	\includegraphics[width=0.95\linewidth]{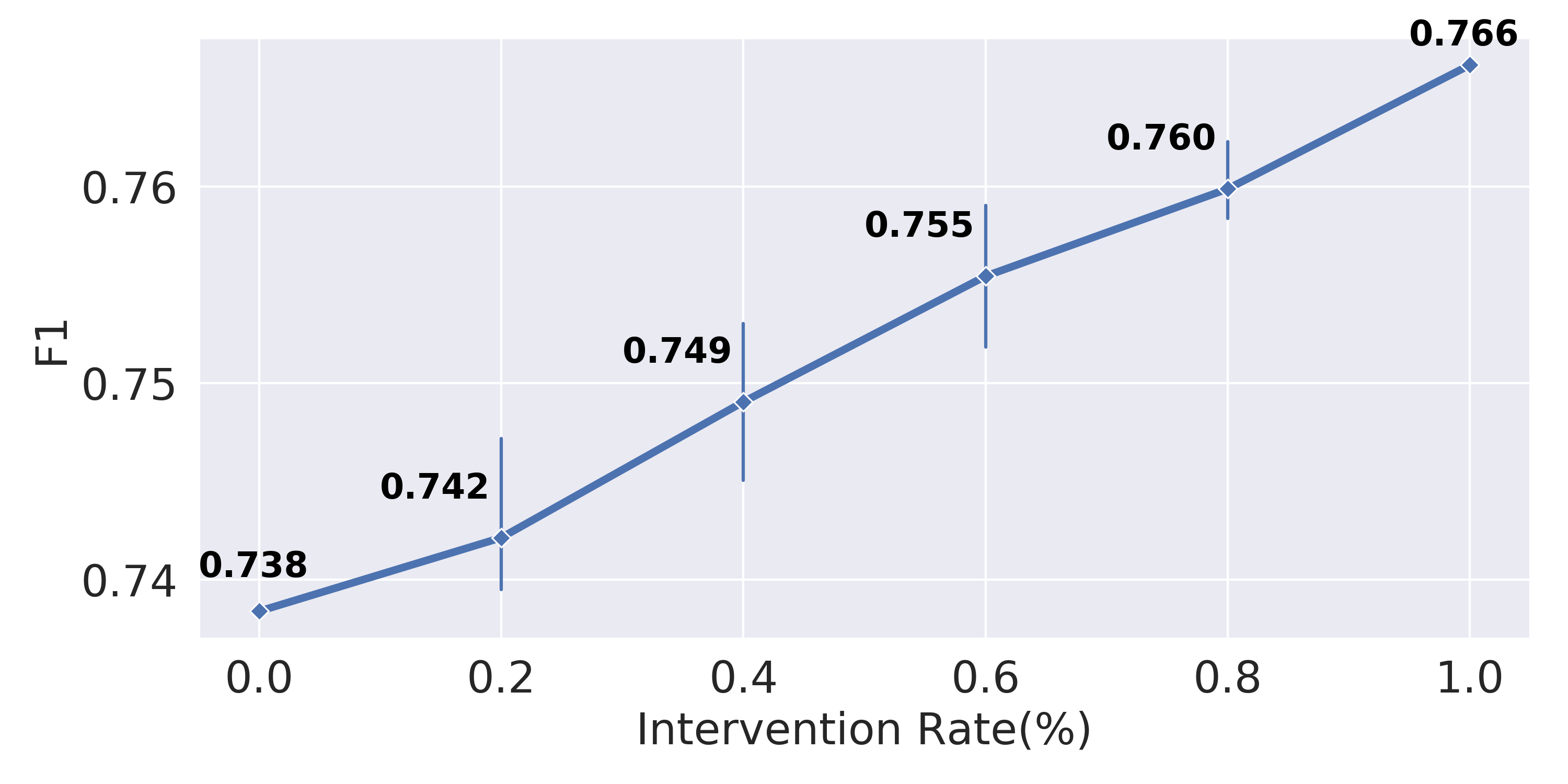}
	\label{fig:dev}
    }
    
    \subfigure[Results of Grasping Success Rate in physical experiment.]{	
    \includegraphics[width=0.95\linewidth]{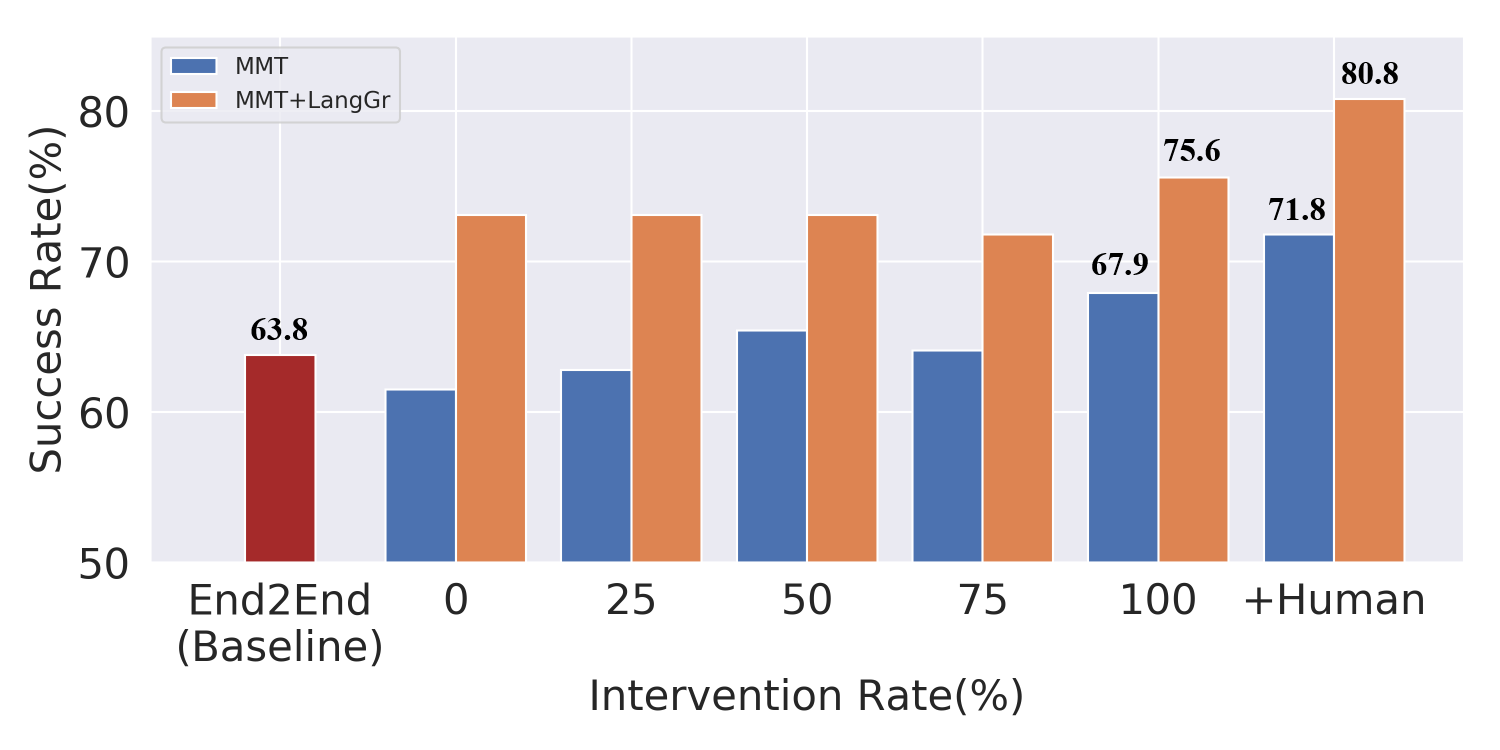}
	\label{fig:exp3}}
	\caption{Evaluation results.}
\end{figure}

\noindent\textbf{Evaluation on Physical Robot} For a final performance test, we conduct extra $80$ grasping trials\footnote{We keep 80 object placements for each setting consistent.} for each model settings corresponding to Table~\ref{table:exp2}. Grasping performance execution by a physical robot is shown in Figure.~\ref{fig:exp3}. 
End2End is the result from our baseline model trained on L-VMRD in Figure.~\ref{fig:model_conf}(a). MMT and MMT+LangGr are models fine-tuning \textbf{image-to-text} or both \textbf{image-to-text and language grounding} respectively in SceneText-BERT setting. \textbf{+human} adds extra human intervention (SceneText-Interv-BERT) with Inter-100 fine-tuning model during grasping.

As shown in Figure.~\ref{fig:exp3}, our proposed method, which fine-tunes on human-intervention collection data, achieves $67.9\%$ and $75.6\%$ success rate compared to our baseline ($63.8\%$). For \textbf{Q3} and \textbf{Q4}, our proposed methods achieve $71.8\% (\uparrow 3.9\%)$ and $80.8\% (\uparrow 5.2\%)$ success rate with human-intervention (\textbf{+human}) compared to without human-intervention ($67.9\%$, $75.6\%$), in which $46.3\%$ scene language description from a robot is incorrect and intervened by human-intervention. This shows that human language intervention can improve the performance of grasping online on a real robot compared to the only image-to-text (self-explanation) method. Moreover, results from \textbf{MMT} and \textbf{MMT+LangGr} show that using human-intervention samples to train models can achieve better performance when there are very few to fine-tune the model. This indicates our proposed human-in-the-loop framework is applicable and performs well on the physical robot.



\section{Related Work}
\noindent\textbf{Natural Language and Robotics} Natural language has been used with a variety of robot platforms, ranging from manipulators to mobile robots to aerial robots \cite{ahn2022can,raychaudhuri2021language,thomason2016learning,chen2021learning,scalise2019improving}. 
Most existing work is related to language understanding and language generation problems. 
For human-to-robot, language grounding is the mainstream means to learn the connection between percepts and actions in visual language navigation \cite{anderson2018vision,ku2020room} and robotic grasping tasks \cite{can2019learning,zellers2021piglet,wang2021ocid}. For robot-to-human, multi-modal natural language generation (NLG) is widely adopted to lessen the communication barriers between humans and robots \cite{vinyals2015show,li2020oscar,cornia2020meshed,yuan2020bridge,shi2021retrieval,zhang2021vinvl}, converting non-verbal data to the language that human can understand \cite{singh2020ai}. For bidirectional human-robot, \citet{yuan2022situ} propose an explainable artificial intelligence system in which a group of robots predicts users' values by taking in situ feedback into consideration while communicating their decision processes to users through explanations.
Our work is in line with the above work in exploiting information from natural language to facilitate decision-making. This is important because natural language is the most intuitive means of human-robot interaction. However, from the aspect of language, the main difference from existing work is that we take a step forward to not only considering language scene description as \textit{input} for a robot but also as an \textit{interface} of the model for online self-explanation simultaneously.


\noindent\textbf{Grasping in Cluttered Scene } Conventional methods \cite{chu2018real,mahler2019learning,morrison2020learning,kumra2020antipodal} focus more on object-agnostic grasp points detection (Figure.~\ref{fig:model_conf} (a)) missing parsing the object stacking scenario. 
For cluttered grasping, scene understanding and human instruction are usually considered.
\citet{zhu2020hierarchical,zhang2019multi} adopt structured scene understanding (e.g., scene graph or relationship tree) to realize cluttered grasping detection.
\citet{mees2020composing,chen2021joint} fuse a natural language command and an observation image to detect a grasping in a two-stage and an end-to-end manner, respectively. \citet{shridhar2020interactive,zhang2021invigorate} receive natural command and image input, and then grasp a specified object. 
Existing work exploits human language to specify an object from the clutter, but does not allow human intervention for error correction. While existing method take language as \textit{external} input, our scene language description is an \textit{internal} component of the model.
Moreover, we adopt the pre-train language model instead of RNN models in existing work.




\section{Conclusion}
We investigate language scene representation to robotic grasping, which enables a robot to explain its object selection to the user and allows the user to intervene with the selection by natural language. Experiments show that the proposed explainable textual scene representation outperforms both object-agnostic and scene-graph based methods. By human language intervention, the performance can be broadly increased. Our results indicate the promise of using NLP models in a robotic system both as a representation and for human intervention. To our knowledge, we are the first to consider textual scene encoding and human correction in robotic grasping tasks, which can improve grasping performance using natural language (vs. w/o human-in-the-loop) and robustness (by pre-trained language model).

\section{Ethical Statement}
Five graduate students who studied electronic engineering are hired to cooperate with a collaborative robot (Kinova) in our real-world experiment. Because of the subject background, they can be easy to decide whether give human language intervention based on human-like grasping behavior each turn. The participants need to annotate the object bounding box for each sample during the data collection stage.   All participants have received labor free corresponding to their amount of trials.

\section*{Acknowledgements}
We would like to thank anonymous reviewers for their insightful comments and suggestions to help improve the paper. This publication has emanated from research conducted with the financial support of the Pioneer and "Leading Goose" R\&D Program of Zhejiang under Grant Number 2022SDXHDX0003, Shanghai Science and Technology Innovation Action Plan under Grant Number 19511120400, and Hangzhou City Agriculture and Social Development General Project under Grant Number 20201203B118. Yue Zhang is the corresponding author.




\bibliography{anthology}
\bibliographystyle{acl_natbib}

\clearpage
\appendix

\section{Appendix}\label{sec:appendix}

\subsection{Details of Dataset}
\subsubsection{Dataset Generation}\label{app:dataset_all}
VMRD contains $4,683$ samples for $31$ classes originally. Each sample has an \textbf{image} labeled with \textbf{object bounding boxes}, \textbf{object class}\footnote{The object class is used to generate language description of the scene and construct the scene graph in our proposed dataset.}, and \textbf{grasp bounding boxes}. Also, stacking relationships between objects in images are provided in a \textbf{relationship tree}. Based on this, we further label various \textbf{language descriptions} and a \textbf{scene graph} for each sample in VMRD\footnote{Note that we filter seven samples with incorrect labeling in the original VMRD.} based on the spatial information from object bounding box and relationship tree. An auxiliary grasp-wise spatial attribute \textbf{surface} is also introduced for better cluttered grasping performance. 

\subsubsection{Dataset Usage}\label{app:dataset_use}
For the image-to-text model, (image and language descriptions) are used. For the language grounding model, (image, language descriptions, and object bounding box) are used. For scene graph generation and graph-based object selection model, (image, scene graph, object bounding box, grasping bounding box) are used. For grasping model (vanilla), (image, object bounding box, grasping bounding box, surface) are used. All of model are setup in Sec.~\ref{sec:model_define}.

\subsubsection{Scene Graph Example}\label{app:sg}
For example, in Figure.\ref{fig:dataset_sg}, the relationship tree only shows relationships as \textit{``mobile phone-on-box''} and \textit{``box-on-notebook''}, but cannot encode the relationship between \textit{``mobile phone''} and \textit{``notebook''} (e.g., \textit{``mobile phone-on-notebook''}).

\subsubsection{Surface Example}\label{app:surface}
In Figure.~\ref{fig:dataset}, the \textit{``notebook''} is stacked by an \textit{``apple''}, thus the \textbf{surface} corresponding to \textit{``apple''} grasping groundtruth is \textit{``False''}. As for the toothpaste, it is not under any other objects, thereby labeled \textit{``True''} for \textbf{surface}. In our task, this attribute can improve grasping performance.

\subsection{Details of Language-based Grasping Model}\label{app:grasping_model}
\subsubsection{Backbone Model}
\begin{figure*}[h]
	\centering 
	\includegraphics[width=0.9\linewidth]{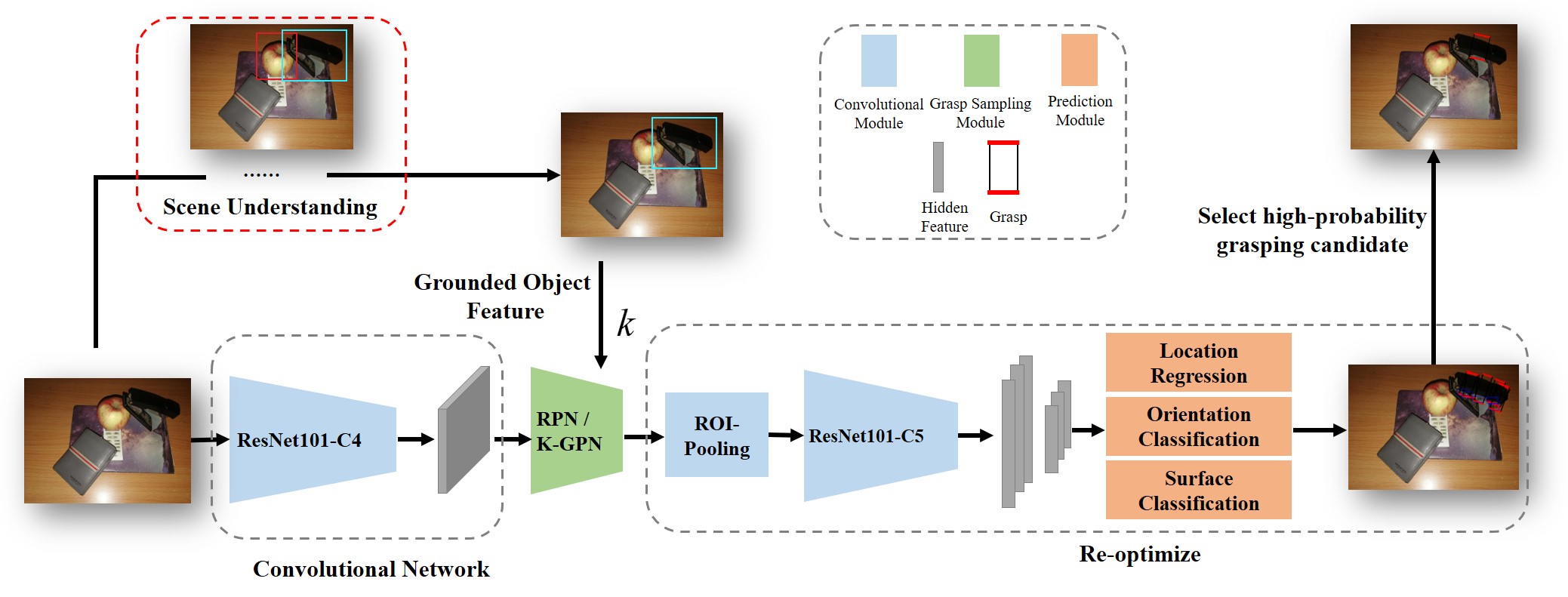}
	\caption{The architecture of Language-based Grasping Model. The input of the scene image is fed into \textbf{ResNet101-C4} to extract visual features, which is one of the inputs of our proposed \textbf{K-GPN}. The grounded object feature $k$ is another input of \textbf{K-GPN}. \textbf{K-GPN} predicts  grasp proposals and output proposal grasping region (also named ROI) features, which forwards into \textbf{ROI-Pooling} and \textbf{ResNet101-C5} to obtain 2048 dimensions feature vectors. These feature vectors are used to predict the final grasping location, orientation, and surface. The highest confidence grasping candidate is selected to execute by a real robot.}
	\label{fig:network_struture2}
\end{figure*}

Our backbone model is developed on top of the two-stage grasp detection pipeline~\cite{chu2018real} (a grasp-version Faster RCNN~\cite{ren2016faster}), fusing the language knowledge as guidance. In doing so, we propose a Knowledge-guided Grasp Proposal Network (K-GPN) to replace with Region Proposal Network (RPN) in Figure.~\ref{fig:network_struture2}, for fusing the grounded object feature with the visual feature. In our framework, we formulate the grasp detection as three parts: (1) Grasp Proposals, (2) Grasp Orientation Classification and Multi-grasp Detection, and (3) Grasp Stacking Classification, described below.

Finally, the highest-confidence angles are selected for each grasp bounding box, and the grasp bounding box (predicted from proposals) corresponding to the highest confidence (mean of the bounding box confidence and surface confidence) is selected as $g_i$ in Eq.~\eqref{eq:grasp_def1} with the selected angle.

\subsubsection{Grasp Proposals}
The module aims to fuse grounded object feature and visual feature to the grasped object. The visual feature used here is a feature map ($z \in {R}^{50 \times 50 \times 1024}$) of the intermediate layers of ResNet-101, and the grounded object feature ($k \in R^{1 \times 4}$, also named $K_g$) is obtained from language grounding model (in Sec.\ref{sec:lang_grounding}) based on \textbf{Self-explanation} or \textbf{Human-intervention}. The proposed K-GPN is employed to fuse $z$ and $k$ and output a new feature vector ${1\times 1 \times 512}$, which is further fed into a two-layer Multi-Layer Perceptron (MLP) to predict the probability of grasp proposal and region of interest (ROI). Different from RPN used by~\cite{chu2018real}, which takes positive and negative proposals with groundtruth over the whole image, the proposed K-GPN samples proposals based on language knowledge and produces ROI related to the \textbf{selected object} in Algorithm~\ref{alg:sample}. The ROI features from \textbf{K-GPN} is fed into the following module to re-optimize and predict the final grasp, shown in Figure.~\ref{fig:network_struture2}. The details are described in Appendix~\ref{sec:orientation_dectection} and \ref{sec:surface_class}.

\begin{algorithm}[tb]
\small
\caption{Knowledge-guided Grasp Proposal Network (\textbf{K-GPN})}
\label{alg:sample}
\textbf{Input}: visual feature $z$, grounded object feature $k$\\
\textbf{Model}: K-GPN\\
\textbf{Output}:Grasp Proposals $G$
\begin{algorithmic}[1] 
\STATE Global proposal set $G_g$ = RPN($z$)
\STATE Positive proposal set and Negative proposal set:$S_p, S_n$
\WHILE{$size(S_p)$ and $size(S_n)$ are less than the sampling count}
\STATE Sampling a grasp proposal $g$ from $G_g$
\IF {$iou(g,g_{gt})>0.5$ and $tiou(g, k)>0.5$}
\STATE Put $g$ into Positive proposal set $S_p$
\ELSE
\STATE Put $g$ into Negative proposal set $S_n$
\ENDIF
\ENDWHILE
\STATE \textbf{return} Grasp Proposals $G=\{S_p, S_n\}$
\end{algorithmic}
\end{algorithm}

As shown in Algorithm~\ref{alg:sample}, our method takes in visual feature $z$ and grounded object feature $k$. And then, the grasp proposals $G$ generated from RPN is selected by satisfying $iou$ constraints and $tiou$ constraints (language knowledge). $g_{gt}$ is the groundtruth grasp corresponding to the proposed $g$. $size(\cdot)$ is the number of the set. $iou(\cdot)$ is the conventional Intersection-over-Union (IoU) function \cite{ren2016faster}. $tiou(\cdot)$ is a function defined in Eq.~\eqref{eq:tiou}, used to select knowledge-guided ROI:

\begin{equation}\label{eq:tiou}
    	tiou\left ( g, k \right )=\frac{\left |g\bigcap k  \right | }{\left |g\right |},
\end{equation}
where $g$ is the grasp proposal, and $k$ is the grounded object feature (visual-language grounded object bounding box).

The proposal loss is defined in Eq.~\eqref{eq:loss_p}
\begin{equation}\label{eq:loss_p}
\begin{aligned}
    {L_p}\left( {\left\{ {\left( {{p_c},{t_c}} \right)_{c = 1}^C} \right\}} \right) &= \sum\limits_c {{L_{cls}}\left( {{p_c},{p^*}} \right)}  \\
    &+ \lambda_1 \sum\limits_c {{p^*}{L_{loc}}\left( {{t_c},{t^*}} \right)},
\end{aligned}
\end{equation}
where $C$ is the set of all proposals, $L_{cls}$ is the cross entropy loss of grasp proposal classification (binary classification). $L_{loc}$ is the smooth L1 regression loss of the proposal locations.
$(p_c, t_c)$ is the binary class and proposal location to the i-th proposal. $p_c$ is True if a grasp is specified, and False if not. $(p^*, t^*)$ is the groundtruth. $\lambda_1$  is the weight coefficient.
 
\subsubsection{Grasp Orientation Classification and Multi-grasp Detection}\label{sec:orientation_dectection}
Our model quantizes the orientation $\theta$ into $R+1$ classification problem by discretizing the continuous orientation angles into $R$ values. Another non-grasp case is also considered in the classification problem and in that case, the grasp proposal is considered incorrect. We select the highest score class as the orientation angle value. In practice, we use equal intervals of $10 ^{\circ}$ to discretize the angles and $R+1=19$. The loss function is as follows:

\begin{equation}\label{eq:loss_g}
\begin{aligned}
{L_g}\left( {\left\{ {\left( {{\rho _l},{\beta _l}} \right)_{c = 0}^C} \right\}} \right) &= \sum\limits_c {{L_{cls}}\left( {{\rho _l},{\rho ^*}} \right)}\\  
&+ \lambda_2 \sum\limits_c {{1_{c \ne 0}(c)}{L_{loc}}\left( {{\beta _c},{\beta ^*}} \right)},
\end{aligned}
\end{equation}
where $\rho_l$ denotes the probability of class $l$ and $\beta_l$ corresponds to the grasp bounding box. $L_{cls}$ is the cross entropy loss of the orientation angle classification (19-class classification). $L_{loc}$ is the smooth L1 loss of bounding boxes with the weight coefficient $\lambda_2$ when angle class $c \ne 0$, where $c = 0$ is short for non-grasp case. $(\rho^*, \beta^*)$ is the groundtruth.

\begin{figure*}[t]
	\centering 
	\includegraphics[width=0.85\linewidth]{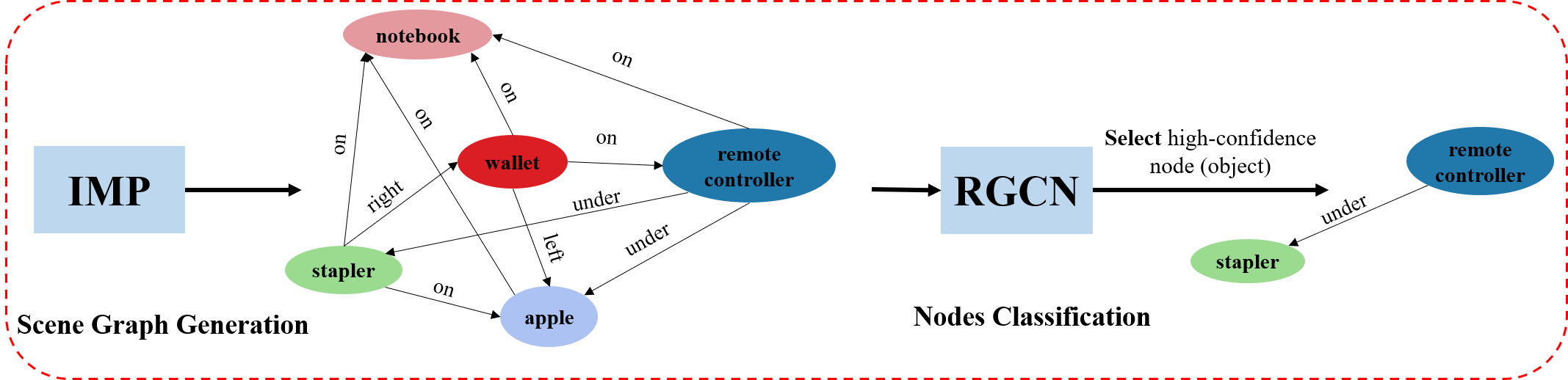}
	\caption{The architecture of scene-graph based scene understanding module.}
	\label{fig:app_sg}
\end{figure*}

\subsubsection{Surface Classification}\label{sec:surface_class}
We propose a binary classification task to predict whether the grasped object is on the top of a stack of objects. The loss function is as follows:
\begin{equation}\label{eq:loss_s}
    {L_s}\left( {\left\{ {\left( {{p_c}} \right)_{c = 1}^C} \right\}} \right) = \sum\limits_c {{L_{cls}}\left( {{s_c},{s^*}} \right)},
\end{equation}
where the same as Eq.~\eqref{eq:loss_p}, $L_{cls}$ is the cross entropy loss of grasp proposal \textbf{surface} classification (binary classification).  $s_c$ is False if the grasped object is stacked by others, and True if not. $s_*$ is the groundtruth. Detail is available in Appendix~\ref{app:surface}.

The total training loss for language-based grasping detection is:

\begin{equation}\label{eq:loss_total}
   L_{total} = L_p + L_g + L_s.
\end{equation}

\subsection{Image-to-text Model}\label{app:image-to-text}
\begin{figure}[t]
	\centering 
	\includegraphics[width=0.8\linewidth]{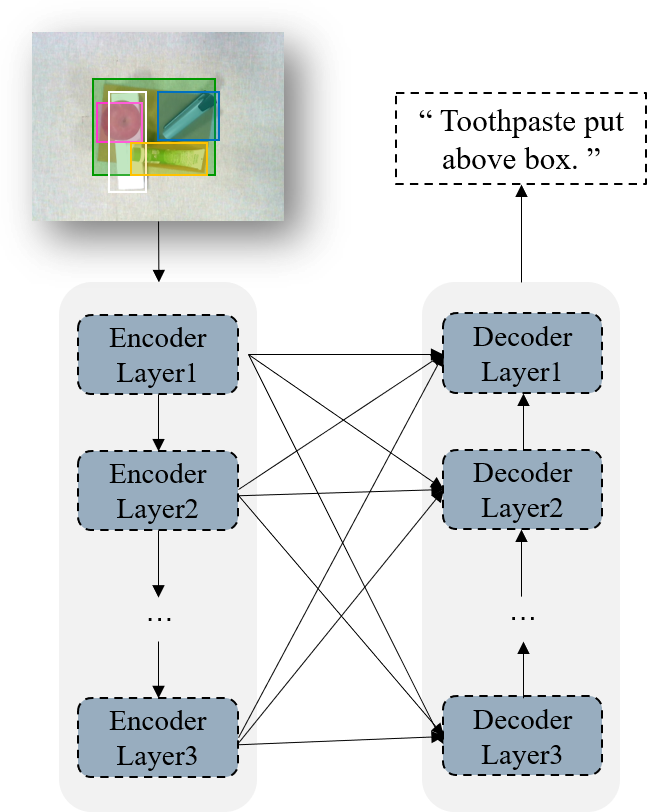}
	\caption{The overview of the image-to-text model applied in our self-explanation pipeline.}
	\label{fig:MMT}
\end{figure}

\textbf{MMT} is shown in Figure.~\ref{fig:MMT}. \textbf{Input} is the region features and bounding box detected from the robot observed image. \textbf{Output} is a description of the spatial relationships of objects in the scene. We hope the subject object can be grasped without collision based on the described spatial relationship.


\subsection{Training Details of Language-based Grasping Model}\label{app:grasp_training}
For the baseline model, during ROI sampling in \textbf{K-GPN}, the positive and negative sampling counts for loss calculations are both $128$. The optimizer is Adam and the learning rate is $1e-4$ for 100 epochs with batch size 8. $\lambda_1$ and $\lambda_2$ are both $1.0$. 

It is noted that for grounded object feature $k$, we first use the groundtruth of the language grounding model to train from scratch and fine-tune the model using the outputs of the language grounding model.

\subsection{Hardware and Software Implementation}\label{app:hard_soft}
The grasping execution is taken place on a single-arm Kinova Jaco 7DOF robot under the framework of Robot Operating System (ROS) Kinetic, shown in Figure.\ref{fig:network_struture}. We use an Intel RealSense SR300 RGB-D camera to obtain RGB-D images mounted on the wrist of the robot. All the computation is completed on a PC running Ubuntu16.04 and Pytorch 1.7 with one Intel Core i7-8700K CPU and one NVIDIA Geforce GTX 1080ti GPU.

\subsection{Metric Details}\label{app:metric}
R@k is the percentage of cases where at least one of the top-k detection is correct. P@k is the correct rate for all top-k predictions.

In the simulation setting, a correctly detected grasp has a Jaccard Index greater than $0.25$ and the absolute orientation error less than $30 ^\circ$ relative to at least one of the groundtruth grasps of the collision-free object \cite{kumra2020antipodal}. 

In the physical setting, for the grounding model, the correct label is Intersection-over-Union (IoU) over 0.5, and for the image-to-text model, the correct label is human-annotated.

\subsection{Scene Graph based method}\label{app:SceneGraph}
\subsubsection{Model Structure}\label{app:sg_model}
The model \textbf{SceneGraph-Rep} uses a scene-graph based scene understanding module (shown in Figure.~\ref{fig:app_sg}) to realize the function of \textcolor{red}{red frame} in Figure.~\ref{fig:network_struture2}.

The module is hierarchical including two sub-modules: \textbf{(i)} Iterative Message Passing (IMP)\cite{xu2017scene} to generate a scene graph in our work. \textbf{(ii)} Relational Graph Convolution Network (RGCN)\cite{schlichtkrull2018modeling} to realize binary classification (can or cannot be grasped for each node). The IMP and RGCN are trained on our proposed L-VMRD same as our proposed method.

The \textbf{input} is region features from common object detection model (Faster RCNN~\cite{ren2016faster}) using scene image $I$. IMP outputs the scene graph $S_g$ in the form of triple (i.e., <subject, predicate, object>), which is fed into RGCN to predict the graspability of each node (object). The high-confidence object is selected corresponding with the bounding box. The whole process can be formulated as follows:
\begin{equation}\label{eq:sg_model}
\begin{aligned}
& Z,B = Detecton\left( I \right),\\
& {S_g} = IMP\left( Z \right),\\
& \left({x_t, y_t, w_t, h_t} \right) = RGCN\left( {{S_g},B} \right),
\end{aligned}
\end{equation}
where $Detecton$ is a common object detection model. $Z$ and $B$ are the set of region features and the set of bounding boxes for each detected object, respectively. $\{x_t, y_t, w_t, h_t\}$ is the bounding box of selected object, same as definition in Sec.\ref{sec:lang_grounding}.


\begin{figure}[t]
	\centering 
	\includegraphics[width=0.9\linewidth]{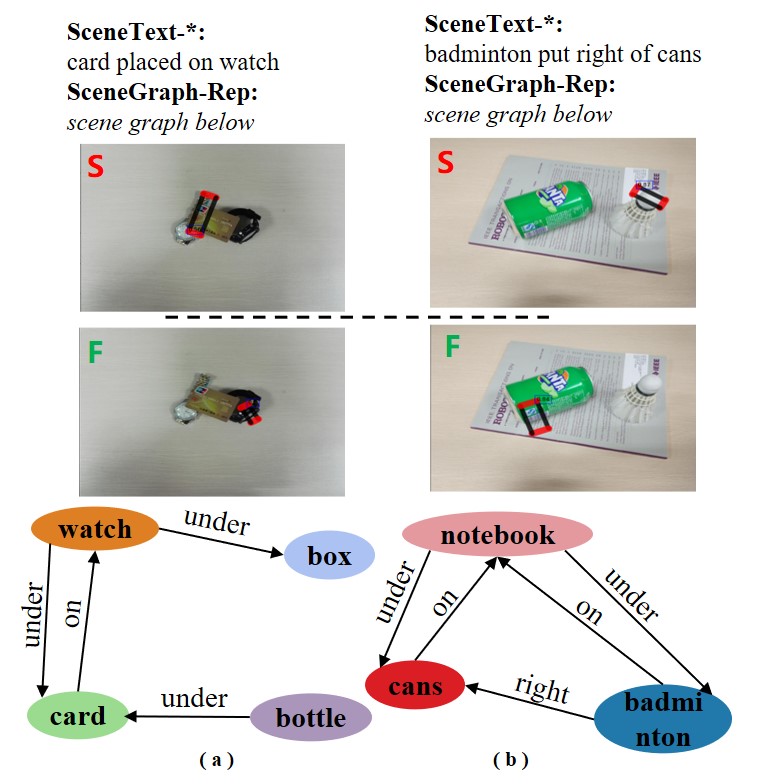}
	\caption{Visualization of MMT (image-to-text) and IMP (scene graph) based self-explanation models. \textcolor{red}{\textbf{S}} means successful case, while \textcolor{green}{\textbf{F}} means failure case.}
	\label{fig:case_study2}
\end{figure}


\subsubsection{Case Study}\label{app:sg_case}
In Figure.~\ref{fig:case_study2}, we give two failure cases caused by low-quality scene graph generation, indicating that \textbf{SceneGraph-Rep} highly depends on the output quality of the scene graph generation model. In Figure.~\ref{fig:case_study2}(a), failure is caused by the incorrect scene graph generation. In Figure.~\ref{fig:case_study2}(b), failure is mainly caused by the error classification from RGCN.




\end{document}